# Sequential Inference for Latent Force Models

**Jouni Hartikainen**
Dept. of Biomedical Engineering
and Computational Science
Aalto University, Finland
jouni.hartikainen@aalto.fi

**Simo Särkkä**
Dept. of Biomedical Engineering
and Computational Science
Aalto University, Finland
simo.sarkka@aalto.fi

## Abstract

Latent force models (LFMs) are hybrid models combining mechanistic principles with non-parametric components. In this article, we shall show how LFMs can be equivalently formulated and solved using the state variable approach. We shall also show how the Gaussian process prior used in LFMs can be equivalently formulated as a linear state-space model driven by a white noise process and how inference on the resulting model can be efficiently implemented using Kalman filter and smoother. Then we shall show how the recently proposed switching LFM can be reformulated using the state variable approach, and how we can construct a probabilistic model for the switches by formulating a similar switching LFM as a switching linear dynamic system (SLDS). We illustrate the performance of the proposed methodology in simulated scenarios and apply it to inferring the switching points in GPS data collected from car movement data in urban environment.

## 1 Introduction

Gaussian process regression [9, 11] refers to the non-parametric Bayesian machine learning approach, where the unknown function $y = x(\mathbf{t})$ is modeled as a Gaussian process (or actually a Gaussian random field). The prediction is done by computing the conditional distribution of the function values $x(\mathbf{t}^*)$ at specific test inputs $\mathbf{t}^*$, where the conditioning is with respect to the training set $\{(\mathbf{t}_i, y_i) : i = 1, \ldots, N\}$. In Gaussian process regression literature (e.g, [11]) the unknown function is usually denoted as $y = f(\mathbf{x})$, where $\mathbf{x}$ is the input and $f(\mathbf{x})$ is the Gaussian process. However, to be consistent with the notation used in stochastic process literature (e.g., [10, 4, 6]), here we shall denote the input as $\mathbf{t}$ and the unknown function as $x(\mathbf{t})$.

When the input variable $t$ is scalar valued, the process $x(t)$ is a Gaussian process in the same sense as in classical analysis of stochastic processes (see, e.g., [10]). As linear operators applied to Gaussian processes result in Gaussian processes, and because solutions of linear differential equations are linear operations on the driving forces, it is also possible to combine linear differential equation models with non-parametric Gaussian process models. The idea of *latent force models* (LFMs) [2, 3] is exactly this and the inference in LFMs is based on computing the various covariance functions between variables by first constructing the explicit solution for each variable in the differential equation, and then computing the covariance functions using these solutions.

Unfortunately, the approach where the differential equations for each variable are solved separately and the covariance functions are constructed from the solutions results in quite tedious closed form computations (cf. [2, 3]), because many of the computations cannot be easily transformed into numerical algorithms. A better approach in this sense is the state variable approach, which is commonly used in Kalman filtering (see, e.g. [4, 6]), which replaces the calculus of scalar differential equations and their impulse responses with solutions of vector valued linear stochastic differential equations driven by Gaussian processes, and their matrix exponential based solutions, that is, matrix valued "impulse responses".

In this article, we shall first show how latent force models can be equivalently formulated and solved using the state variable approach. In the reformulation the theoretical solutions itself remain the same, but the implementation can be reduced to computation of matrix exponentials and their integrals, which are easier to implement numerically.

As shown in [7], a time-series type of Gaussian process prior can be equivalently formulated as a state-space model, that is, as a linear system driven by a white noise process. This means that the latent Gaussian processes used in LFM models can be implemented by extending the state-space model with additional state variables representing the latent force components. In this article we shall follow this approach and reformulate LFMs as linear state-space models driven by white noise processes. As the resulting model is a Gauss-Markov model, we shall also show how Kalman filter and smoother (cf. [7]) can be used for doing efficient inference on the resulting model.

We shall also show how the switching LFM [3] can be equivalently reformulated using the state variable approach. We briefly discuss some of the problems with the switching LFMs, and propose an alternative approach, in which the state-space view on LFMs is utilized to perform probabilistic inference for the switching sequence. This reformulation leads to switching linear dynamic systems, which can be efficiently treated with classical multiple model approaches [4] or by using more recently developed methodology [5].

We illustrate the performance of the proposed methodology in simulated scenarios, and as a real world data example apply the constructed switching LFM for inferring the switching points in GPS data collected from car movement data in urban environment.

## 2 Latent Force Models

In [2] Alvarez et al. introduced *latent force models* (LFMs), which are hybrid models combining mechanistic principles with non-parametric components. For example, in [2] $D$ output processes $\{x_d(t)\}_{d=1}^D$ are modelled as second order differential equations

$$A_d \frac{d^2 x_d(t)}{dt^2} + C_d \frac{x_d(t)}{dt} + \kappa_d x_d(t) = \sum_{r=1}^R S_{d,r} u_r(t), \quad (1)$$

where the driving processes $u_r(t)$ are given independent Gaussian process (GP) [11] priors $u_r(t) \sim \mathcal{GP}(m(t), k_{u_r}(t, t'))$, $r = 1, \ldots, R$ where $m(t)$ is an appropriate mean function (taken usually to be zero without loss of generality) and $k_{u_r}(t, t')$ a suitably chosen covariance function.

The inference in the approach of Alvarez et al. [2] is based on closed form computation of the covariance functions of $x_d(t)$, $dx_d(t)/dt$ and all the required cross covariances by solving the differential equation. The derivations in [2] were done in purely scalar notation. An alternative approach would be to work with vectors and matrices and convert the output model into state-space model, which in case of second order model (1) can be done as follows:

1. Define state and input vectors as $\mathbf{x}(t) = (x_1(t)\, dx_1(t)/dt\, \ldots\, x_D(t)\, dx_D(t)/dt)^T$ and $\mathbf{u}(t) = (u_1(t)\, \ldots\, u_R(t))^T$.

2. Define matrices

$$\mathbf{F} = \begin{pmatrix} 0 & 1 & & & \\ -\frac{\kappa_1}{A_1} & -\frac{C_1}{A_1} & & & \\ & & \ddots & & \\ & & & 0 & 1 \\ & & & -\frac{\kappa_D}{A_D} & -\frac{C_D}{A_D} \end{pmatrix} \quad (2)$$

and

$$\mathbf{L} = \begin{pmatrix} 0 & \cdots & 0 \\ -\frac{S_{1,1}}{A_1} & \cdots & -\frac{S_{1,R}}{A_1} \\ \vdots & \ddots & \\ 0 & & 0 \\ -\frac{S_{D,1}}{A_D} & & -\frac{S_{D,R}}{A_D} \end{pmatrix}. \quad (3)$$

Now the model can be written in form

$$\frac{d\mathbf{x}(t)}{dt} = \mathbf{F}\mathbf{x}(t) + \mathbf{L}\mathbf{u}(t). \quad (4)$$

The differential equation has the solution

$$\mathbf{x}(t) = \Phi(t)\mathbf{x}(t_0) + \int_{t_0}^t \Phi(t-s)\mathbf{L}\mathbf{u}(s)ds, \quad (5)$$

where $\Phi(\tau)$ denotes the matrix exponential $\Phi(\tau) = \exp(\mathbf{F}\,\tau)$. In this case it happens that the matrix exponential can be easily computed in closed form. All the required covariance terms could now be evaluated as follows:

$$\begin{aligned} E[\mathbf{x}(t)\mathbf{x}(t')] &= \Phi(t-t_0)\mathbf{P}_x^0 \Phi(t'-t_0)^T \\ &+ \int_{t_0}^{t'}\int_{t_0}^{t} \Phi(t-s)\mathbf{L}\mathbf{K}_{uu}(s,s')\mathbf{L}^T \Phi(t'-s')^T ds ds', \end{aligned} \quad (6)$$

where $\mathbf{P}_x^0$ is the prior covariance of $\mathbf{x}(t)$ and $\mathbf{K}_{uu}(s, s')$ is the joint covariance of all the latent forces between time instants $s$ and $s'$. Since we assume independence across forces, $\mathbf{K}_{uu}(s, s')$ is diagonal. The difficulty here is how to evaluate the double integral in (6). If the covariance functions of the latent forces are set to squared exponentials

$$k_{u_r}(\tau) = \exp\left(-\frac{\tau^2}{l_r^2}\right), \quad \tau = t - t', \ r = 1 \ldots R, \quad (7)$$

the covariance functions $k_{y_i, x_j}(t, t')$, $k_{x_i, x_j}(t, t')$, $k_{x_i, u_r}(t, t')$ and $k_{y_i, u_r}(t, t')$ can be solved analytically for certain output models, such as (1). This enables the usage of standard GP regression techniques for predicting the values of $\mathbf{x}(t)$ and $\mathbf{u}(t)$ in arbitrary time points as well as for evaluating the marginal data likelihood $p(\mathbf{y}|\theta) = \int p(\mathbf{y}|\mathbf{x},\theta)p(\mathbf{x}|\theta)d\mathbf{x}$, where $\theta$ contains the parameters of output model (4) and the covariance functions $k_{u_r}$.

## 2.1 Sequential Gaussian Process Priors for Latent Forces

A drawback of the direct GP regression solution is that the computational complexity scales as $\mathcal{O}(D^3 T^3)$, where $T$ is the number of time instances in the observations. In [2] multioutput generalization of sparse approximations were used to reduce this to $\mathcal{O}(DTK^2)$, where $K$ is the number of inducing variables used in representing $\mathbf{u}(t)$. While at first glance this scaling appears to be linear in $T$, we argue that this isn't the case in practice since $K$ needs to be increased when $T$ increases so that the data can be modelled appropriately. Perhaps an even more severe difficulty with the direct GP solution is that one has to always solve the needed covariance functions when constructing new output models. This can be very challenging or even impossible in many cases, and thereby imposes serious restrictions on the generality of the modelling framework.

To remedy these problems we propose to use the techniques presented in [7] for formulating the GP priors on the components $r = 1, \ldots, R$ of $\mathbf{u}(t)$ as a multivariate linear time-invariant (LTI) stochastic differential equation (SDE) models of form

$$\frac{d\mathbf{z}_r(t)}{dt} = \mathbf{F}_{z,r}\,\mathbf{z}_r(t) + \mathbf{L}_{z,r}\,w_{z,r}(t) \qquad (8)$$

where $\mathbf{z}_r(t) = (u_r(t)\ \frac{du_r(t)}{dt}\ \cdots\ \frac{d^{d_r-1}u_r(t)}{dt^{d_r-1}})^T$ and

$$\mathbf{F}_{z,r} = \begin{pmatrix} 0 & 1 & & \\ & \ddots & \ddots & \\ & & 0 & 1 \\ -a_r^0 & \cdots & -a_r^{p_r-2} & -a_r^{p_r-1} \end{pmatrix}, \mathbf{L}_{z,r} = \begin{pmatrix} 0 \\ \vdots \\ 0 \\ 1 \end{pmatrix}.$$

By choosing the coefficients $a_r^0, \ldots, a_r^{p_r-1}$, the spectral density $q_r$ of white noise process $w_{z,r}(t)$ and the dimensionality $p_r$ of $\mathbf{z}_r(t)$ appropriately the dynamic model on $u_r(t)$ can be chosen to correspond a GP prior with a certain stationary covariance function. We are especially interested in covariance functions of form

$$k_{u_r}(\tau) = \exp\left(-\frac{\sqrt{2(p_r+1/2)}\tau}{l}\right) \frac{\Gamma(p_r+1)}{\Gamma(2p_r+1)}$$
$$\times \sum_{i=0}^{p_r} \frac{(p_r+i)!}{i!(p_r-i)!}\left(\frac{\sqrt{8(p_r+1/2)}\tau}{l}\right)^{p_r-i},$$

which is the Matérn class of covariance functions with smoothess parameter $\nu = p_r + 1/2$. This class is particularly useful since it contains the exponential and squared exponential covariances as special cases ($\nu = 1/2$ and $\nu \to \infty$). The key property of this model class is that it has an analytic state-space representation since its spectral density $S(\omega)$ can be written as a rational function of $\omega^2$ [7]. If one wishes to use the squared exponential covariance function (which has no analytic Gauss-Markov representation, since it is infinitely differentiable) instead, more quickly converging state-space representations can be constructed by applying Taylor series approximations for the spectral density [7].

The GP prior models of form (8) can be straightforwardly augmented to output model (4) to form a joint model

$$\frac{d\mathbf{x}_a(t)}{dt} = \mathbf{F}_a \mathbf{x}_a(t) + \mathbf{L}_a\,\mathbf{w}_a(t) \qquad (9)$$

where we have defined an augmented the state vector $\mathbf{x}_a(t) = (\mathbf{x}(t)^T\ \mathbf{z}_1(t)^T\ \cdots\ \mathbf{z}_R(t)^T)^T$, and the matrices $\mathbf{F}_a$ and $\mathbf{L}_a$ are constructed such that they operate on the augmented state appropriately. As an example consider the second order latent force model (1) with $D = R = 1$ and $p_1 = 2$, in which case the state vector of the joint model is $\mathbf{x}_a(t) = (x_1(t)\ \frac{dx_1(t)}{dt}\ u_1(t)\ \frac{du_1(t)}{dt})^T$ and the dynamic model matrices are

$$\mathbf{F}_a = \begin{pmatrix} 0 & 1 & 0 & 0 \\ -\frac{\kappa_1}{A_1} & -\frac{C_1}{A_1} & -\frac{S_{1,1}}{A_1} & 0 \\ 0 & 0 & 0 & 1 \\ 0 & 0 & -a_1^0 & -a_1^1 \end{pmatrix}, \mathbf{L}_a = \begin{pmatrix} 0 \\ 0 \\ 0 \\ 1 \end{pmatrix}.$$

Higher dimensional models can be constructed in a similar fashion.

## 2.2 Posterior Inference and Predictions

The LTI SDE model (9) has the fortunate property that it can be analytically converted to a discrete-time dynamic model

$$\mathbf{x}_k = \mathbf{A}(\Delta t_k)\mathbf{x}_{k-1} + \mathbf{q}_{k-1},\ \mathbf{q}_{k-1} \sim N(\mathbf{0}, \mathbf{Q}(\Delta t_k)), \qquad (10)$$

where the transition and process noise matrices can be solved on the time instances $\mathcal{T} = \{t_k\}_{k=1}^T$ as

$$\mathbf{A}(\Delta t_k) = \Phi_a(\Delta t_k), \Delta t_k = t_k - t_{k-1}, \Phi_a(\tau) = \exp(\mathbf{F}_a\,\tau),$$
$$\mathbf{Q}(\Delta t_k) = \int_0^{\Delta t_k} \Phi_a(\Delta t_k - \tau)\,\mathbf{L}_a\,\mathbf{Q}_c\,\mathbf{L}_a^T\,\Phi_a(\Delta t_k - \tau)^T\,\mathrm{d}\tau, \qquad (11)$$

where $\mathbf{Q}_c$ is the spectral density of white noise process $\mathbf{w}_a(t)$ in (9). So far we have not discussed how the

output process is observed. The standard approach is to use the linear-Gaussian model

$$\mathbf{y}_k = \mathbf{H}_k \mathbf{x}_k + \mathbf{r}_k, \ \mathbf{r}_k \sim N(\mathbf{0}, \mathbf{R}_k), \quad (12)$$

where the matrix $\mathbf{H}_k$ collects the observed components from the state vector.

The filtered posterior distribution of the state $p(\mathbf{x}_k|\mathbf{y}_{1:k}, \theta) = N(\mathbf{m}_k, \mathbf{P}_k)$ on the selected time points can be solved exactly with the classical Kalman filter and the smoothing distribution $p(\mathbf{x}_k|\mathbf{y}_{1:T}, \theta) = N(\tilde{\mathbf{m}}_k, \tilde{\mathbf{P}}_k)$ with the Rauch-Tung-Striebel (RTS) smoother (see, e.g, [4, 6]). Both the Kalman filter and RTS smoother scale in $\mathcal{O}(d^3 T)$ computations, where $d$ is the dimensionality of $\mathbf{x}$ and $T$ the number of time points. The estimation should be started from the Gaussian prior $p(\mathbf{x}_0|\theta) = N(\mathbf{m}^0, \mathbf{P}^0)$, where it is reasonable to set the covariance matrix to be block diagonal of form $\mathbf{P}^0 = \text{blkdiag}(\mathbf{P}^0_x, \mathbf{P}^0_{u_1}, \ldots, \mathbf{P}^0_{u_R})$, where $\mathbf{P}^0_x$ is the joint prior covariance for the non-augmented output process $\mathbf{x}(t)$ chosen according to a priori knowledge. The blocks $\mathbf{P}^0_{u_r}$ for the $R$ latent forces can be set to stationary covariances by numerically solving the algebraic Riccati equations

$$\frac{d\mathbf{P}_{u_r}}{dt} = \mathbf{F}_{z,r} \mathbf{P}_{u_r} + \mathbf{P}_{u_r} \mathbf{F}_{z,r}^T + \mathbf{L}_{z,r} q_r \mathbf{L}_{z,r}^T = \mathbf{0}. \quad (13)$$

Suppose we wish to estimate the smoothing distribution of the state on unobserved time instant $t_*$, that is, $p(\mathbf{x}(t_*)|\mathbf{y}_{1:T}, \theta) = N(\mathbf{m}^s_*, \mathbf{P}^s_*)$, where $t_{k-1} < t_* < t_k$. This can be done by adding $t_*$ to the set of selected time steps $\mathcal{T}$ and skipping the Kalman filter update step on that particular time step on the forward pass, and then running the RTS smoother on $\mathcal{T} \cup \{t_*\}$. Alternatively, after running the Kalman filter and smoother on time steps $\mathcal{T}$ one can infer the state on $t_*$ by first making a prediction on $t_*$ from $p(\mathbf{x}_{k-1}|\mathbf{y}_{1:k-1}, \theta)$ and then smoothing the estimate with $p(\mathbf{x}_k|\mathbf{y}_{1:T}, \theta)$.

In case of non-linear observation models the state trajectory is analytically intractable, but a wide range of Gaussian filters [8, 13] and smoothers [12] has been proposed in the literature.

## 3 Switched Latent Force Models

In [3] the LFM framework was extended to a case of having a system, in which the driving latent forces can switch on certain time instants. The switching process was formulated such that time series was divided into non-overlapping intervals $[t_{q-1}, t_q]_{q=1}^Q$ in which only one latent force $u_{q-1}(t)$ out of $Q$ independent forces $\{u_q(t)\}_{q=1}^Q$ is active one at a time.

In essence, the contribution of paper [3] is to solve (6) for output model (1) analytically in cases when the covariance function of the GP prior for the latent force is

$$k_{uu}(t,t') = \begin{cases} k_{q(t)}(t,t'), & \text{if } q(t) = q(t'), \\ 0, & \text{otherwise}, \end{cases} \quad (14)$$

where $q(t)$ returns the segment index of time point $t$. In [3] the derivation was done in scalar notation, but here we briefly give an alternative derivation in vector form. Assume now that $t \in [t_{\hat{q}-1}, t_{\hat{q}}]$ and $t' \in [t_{\hat{q}'-1}, t_{\hat{q}'}]$ with $Q_m = \min(\hat{q}, \hat{q}')$. Denote $\Psi(t, s, t', s', k(\cdot, \cdot)) = \Phi(t-s)\mathbf{L}k(s,s')\mathbf{L}^T\Phi(t'-s')^T$. In this case the equation (6) still holds and the double integral in it can be solved as

$$\int_{t_0}^{t'} \int_{t_0}^{t} \Psi(t,s,t',s',k_{uu})dsds',$$

$$= \sum_{q=1}^{Q_m-1} \iiint_{t_{q-1}}^{t_q} \Psi(t,s,t',s',k_q)dsds'$$

$$+ \int_{t_{Q_m-1}}^{t'} \int_{t_{Q_m-1}}^{t} \Psi(t,s,t',s',k_{Q_m})dsds',$$

$$= \sum_{q=1}^{Q} \Phi(t-t_q) \iint_0^{\Delta t_q} \Psi(\Delta t_q, s, \Delta t_q, s', k_q)dsds' \Phi(t'-t_q)$$

$$+ \iint_{t_{Q_m-1}}^{t_{Q_m}} \Psi(t,s,t',s',k_{Q_m})dsds',$$

$$= \sum_{q=1}^{Q} \Phi(t-t_q)\mathbf{K}^q_{xx}(\Delta t_q, \Delta t_q)\Phi(t'-t_q)^T,$$

$$+ \Phi(t-t_{Q_m-1})\mathbf{K}^{Q_m}_{xx}(\Delta t_{Q_m}, \Delta t'_{Q_m})\Phi(t'-t_{Q_m-1}),$$

where $\Delta t_q = t_q - t_{q-1}$, $\Delta t_{Q_m} = \min(t_{\hat{q}}, t_{Q_m}) - t_{Q_m-1}$ and $\Delta t'_{Q_m} = \min(t_{\hat{q}'}, t_{Q_m}) - t_{Q_m-1}$. The key here is to note that integral of $\Psi$ with respect to $s$ and $s'$ depends only on the lengths of the integral limits, and thus we can translate the limits above.

### 3.1 Sequential Gaussian Process Priors

As with the standard LFMs, the sequential GP priors of form (8) can also be straightforwardly incorporated to switched latent force models if the switching points are assumed to be known. Inference can be done with a Kalman filter and smoother such that the switching points $\{t_q\}_{q=1}^{Q-1}$ are included to set of time points $\mathcal{T}$, and when making the Kalman filter prediction step to a swithing point $q$ the transition and process noise matrices are set to $\mathbf{A} = \text{blkdiag}(\mathbf{A}_x, \mathbf{0}_p)$ and $\mathbf{Q} = \text{blkdiag}(\mathbf{Q}_x, \mathbf{P}^0_{u_q})$, where $t_k$ is the point before $t_q$ in $\mathcal{T}$ and $p$ the dimensionality of the GP prior that is common to all $Q$ forces. The matrices for the output components $\mathbf{A}_x$ and $\mathbf{Q}_x$ are solved similarly as in (11) with $\mathbf{F}_a = \mathbf{F}$, $\mathbf{L}_a = \mathbf{L}$ and $\Delta t_k = t_k - t_q$.

### 3.2 Probabilistic Model for Switches

In [3] the switching points $\{t_q\}_{q=1}^{Q-1}$ were treated as hyperparameters of the constructed covariance function, which were then optimized with respect to marginal likelihood alongside with the other hyperparameters of the model. While this can be sensible in some cases, placing the swithing points incorrectly can result in erroneous results since it ignores all the uncertainty related to locations of the points. Moreover, the number of segments $Q$ was fixed in [3], and estimated with computationally demanding cross-validation.

To make progress on this we can utilize the state-space view of LFMs by formulating the latent force model as a switching linear dynamical system (SLDS) of form

$$p(\mathbf{x}_k|\mathbf{x}_{k-1}, s_k) = N(\mathbf{x}_k|\mathbf{A}(\Delta t_k, s_k)\mathbf{x}_{k-1}, \mathbf{Q}(\Delta t_k, s_k))$$
$$p(\mathbf{y}_k|\mathbf{x}_k, s_k) = N(\mathbf{y}_k|\mathbf{H}(s_k)\mathbf{x}_{k-1}, \mathbf{R}(s_k)) \quad (15)$$

where $s_k$ denotes the active model at time index $k$. For simplicity we assume a discrete-time Markov model of form $p(s_k|s_{k-1})$ for the model transitions over finite time steps, but we could also alternatively first formulate a continous-time Markov process for the state and model transitions, and then discretize the model on time steps of interest. The key advantage of this approach is that it allows to make probabilistic inference over the model sequence by utilizing the state-of-the-art methodology for SLDSs that are discussed in the next section.

For the SLDS we consider here we assume that there are $R$ active forces on each time step (recall that in [3] only one force was allowed to be active), of which each $r$th force can have $M_r$ different length-scales. To achieve discontinuities in the latent forces similarly as with model of previous section, we assume that the transitions between the different models can happen only via resetting models, which reset the latent force components of the state vector to a suitable prior while keeping the output components intact. For simplicity in this work we assume that the resetting model resets all the latent force components to a zero-mean Gaussian prior with a suitably chosen covariance $\tilde{\mathbf{P}}_u^0 = \text{blkdiag}(\tilde{\mathbf{P}}_{u_1}^0, \ldots, \tilde{\mathbf{P}}_{u_R}^0)$. Thus, the matrices $\mathbf{A}_k$ and $\mathbf{Q}_k$ for the reset model can be implemented as $\mathbf{A}_k = \text{blkdiag}(\mathbf{A}_x, \mathbf{0}_p)$ and $\mathbf{Q}_k = \text{blkdiag}(\mathbf{Q}_x, \tilde{\mathbf{P}}_u^0)$, where $p = \sum_{r=1}^R p_r$, and $\mathbf{A}_x$ and $\mathbf{Q}_x$ are solved similarly as in (11) with $\mathbf{F}_a = \mathbf{F}$, $\mathbf{L}_a = \mathbf{L}$ and $\Delta t_k = t_k - t_{k-1}$.

We also assume that there are $L$ possible length-scales that are shared between the $R$ forces, that is, there are a total of $L^R + 1$ models in the SLDS, of which the $L^R+1$th model is the reset model. The Markov model for the transitions can be stated as

$$p(s_k|s_{k-1}) = \begin{cases} a_{s_k}, & \text{if } s_k = s_{k-1}, \\ b_{s_k}, & \text{if } s_k = L^R + 1 \text{ and } s_{k-1} \neq L^R + 1, \\ c_{s_k}, & \text{if } s_k \neq L^R + 1 \text{ and } s_{k-1} = L^R + 1, \\ 0, & \text{otherwise,} \end{cases}$$

where we require that $a_{s_k} + b_{s_k} = 1$ and $\sum_{s_k=1}^{L^R} c_{s_k} = 1$.

In the switching model presented in [3] the covariance function hyperparameters were considered to be different for different time segments, which results in $Q$ length-scale hyperparameters to be estimated from data if one does not fix the parameters to be the same across the segments. In our SLDS model we can have $L \leq Q$ parameters to be learned, which can be used for more than one segment in the time series, while in the approach of [3] each of the parameters are only used within a single segment.

### 3.3 Inference in Switching Linear Dynamic Systems

Assume now that we have a switching linear dynamic system (15) for the state trajectory and observations, and a Markov model $p(s_k|s_{k-1})$ for the model transitions. It is well known that analytic inference on this class of models scales exponentially with respect to $T$ [4], making it computationally infeasible. Thus, we need to turn to approximations in practical computations.

We consider applying Gaussian sum filtering and smoothing, in which the forward pass is usually termed as *assumed density filtering* (ADF). There are many ways to perform the smoothing pass, but in particular we focus on employing the *expectation correction* (EC) algorithm [5], which can be seen as an analog of applying RTS smoothing to ADF in a similar way as regular RTS smoother is applied to Kalman filtering.

The result of ADF on time step $k$ is a Gaussian mixture

$$p(\mathbf{x}_k|s_k, \mathbf{y}_{1:k}) \approx \sum_{i=1}^{I} w_{i,s_k,k} N(\mathbf{x}_k|\mathbf{m}_{i,s_k,k}, \mathbf{P}_{i,s_k,k})$$

for the state of each model $s_k$ and an approximation for the model probabilities $p(s_k|\mathbf{y}_{1:k})$. The number of mixture components $I$ is chosen according to computational budget (by setting $I_k = M^k$ the result is exact). One step of ADF requires running $IM^2$ Kalman filters, resulting in overall complexity of $\mathcal{O}(d^3 IM^2 T)$.

Similarly, EC produces a mixture approximation for the smoothed distribution as

$$p(\mathbf{x}_k|s_k, \mathbf{y}_{1:T}) \approx \sum_{i=1}^{J} \tilde{w}_{i,s_k,k} N(\mathbf{x}_k|\tilde{\mathbf{m}}_{i,s_k,k}, \tilde{\mathbf{P}}_{i,s_k,k}),$$

and an approximation for $p(s_k|\mathbf{y}_{1:T})$. Similarly as with ADF the number of mixture components $J$ is chosen according to available computational resources. EC is computationally more demanding than ADF, requiring $IJM^2$ RTS smoothers to be run on each step, which results in $\mathcal{O}(d^3 IJM^2 T)$ overall complexity.

For complete details about ADF and EC we refer the reader to [5], but there are few implementation details that need to be discussed briefly. In ADF and EC one needs to collapse a Gaussian mixture of $N$ components to a smaller mixture of $K < N$ components (in ADF $N = IM$ and $K = I$, and in EC $N = IJM$ and $K = J$). There are many ways to do this, but we implemented the same procedure as in [5], where $K - 1$ components are retained directly and the remaining $N - K$ components are merged together via moment matching. For the models we consider here we observed this to be working well. Additionally, in EC one needs to evaluate integrals of form

$$p(s_k|s_{k+1}, \mathbf{y}_{1:T}) = \int p(s_k|\mathbf{x}_{k+1}, s_{k+1}, \mathbf{y}_{1:k}) \\ \times p(\mathbf{x}_{k+1}|s_{k+1}, \mathbf{y}_{1:T}) d\mathbf{x}_{k+1}. \quad (16)$$

We tested approximating this with a numerical Cubature, but that turned out to result in worse overall estimation accuracy than simply evaluating the integrand of (16) in the mean of $p(\mathbf{x}_{k+1}|s_{k+1}, \mathbf{y}_{1:T})$, which was also the way of approximating (16) in [5].

## 4 Experiments

### 4.1 Comparison of Computational Efficiency

In this experiment we compare the computational efficiency and estimation accuracy of the proposed state-space approach to standard LFMs to previously presented methodology, that is, multioutput generalizations of FITC and PITC sparse approximations [1] as well as standard full GPs. We consider a simple scenario, in which we have a second order output model (1) with $D = 5$ outputs and $R = 1$ latent force. We used unit values for all parameters of the output model and covariance functions with the exception that we deviated the sensitivities $S_{d,r}$ slightly from unity. We generated data sets of length $T = 100$ and $T = 500$ on the unit interval and calculated the RMSE values of estimating the output process with all the tested methods. The hyperparameters were fixed to their true values. For FITC and PITC we used 40 inducing inputs placed regularly over the unit interval. The results are reported in Table 1. It can be seen that in terms of RMSE values all methods give comparable performance. We also note that decreasing the length-scale and/or the number of inducing variables will cause the sparse approximations to eventually fail when the underlying latent function is not smooth enough. The state-space approach does not have this problem. We also encountered some numerical difficulties when implementing the needed covariance functions for $x_d(t)$ that were not present in the state-space model.

Table 1: **Comparison of RMSE and CPU time in the simulated example.** The RMSE values are calculate as means over 100 simulations and multiplied by 100. Here KF-G and KF-M refer to state-space LFMs, where KF-G has the squared exponential covariance function (Taylor series approximated with 6 state components [7]) and KF-M the Matérn covariance ($\nu = 3/2$).

|            | GP    | FITC  | PITC   | KF-G  | KF-M  |
|------------|-------|-------|--------|-------|-------|
| **T = 100** |       |       |        |       |       |
| RMSE       | 1.64  | 1.65  | 1.64   | 1.64  | 1.65  |
| CPU (s)    | 0.092 | 0.006 | 0.0291 | 0.024 | 0.021 |
| **T = 500** |       |       |        |       |       |
| RMSE       | 0.82  | 0.83  | 0.82   | 0.82  | 0.82  |
| CPU (s)    | 9.683 | 0.155 | 2.802  | 0.121 | 0.105 |

We also computed the CPU times needed in the inference. For all GPs the recorded times include only the time taken to computate of posterior mean for $x_d(t)$ with precomputed covariance matrices while in state-space models the times include also the calculation of marginal posterior covariances for both $x_d(t)$ and $u_r(t)$ as well as the evaluation of marginal likelihood, which are computed always during the Kalman filter and smoother. In terms of CPU time FITC is the fastest with $T = 100$ data points, but with $T = 500$ and more data points Kalman filtering and smoothing begins to gain the edge in speed.

### 4.2 Performance of the Probabilistic SLFM

In this toy example we illustrate the performance of ADF and EC in estimating the proposed switching LFM. We generate data from a model (1) with $D = 3$ outputs and $R = 1$ latent force (Matérn covariance with $\nu = 3/2$), which had $L = 2$ possible lengthscales, $l_1 = 2$ and $l_2 = 30$. The constants in the Markov transition model were set to $a_1 = a_2 = 0.98$ and $c_1 = c_2 = 0.5$. In the output model we used the parameters $A_1 = A_2 = A_3 = 0.1$, $C_1 = 2$, $C_2 = 3$, $C_3 = 0.5$, $\kappa_1 = 0.4$, $\kappa_2 = 1$, $\kappa_3 = 1$, $S_{1,1} = S_{3,1} = 1$ and $S_{1,2} = 5$. Figure 1 shows a typical result of estimating $x_1(t)$, $u(t)$ and model transitions with ADF and EC with both $I = J = 1$ and $I = J = 3$ mixture components. For reference Kalman filtering and smoothing results with the true transitions are also shown. It can be seen that with one Gaussian com-ponent EC already provides credible results, and the usage of a mixture approximation corrects the length-scale estimate around the mid-right part of the time series.

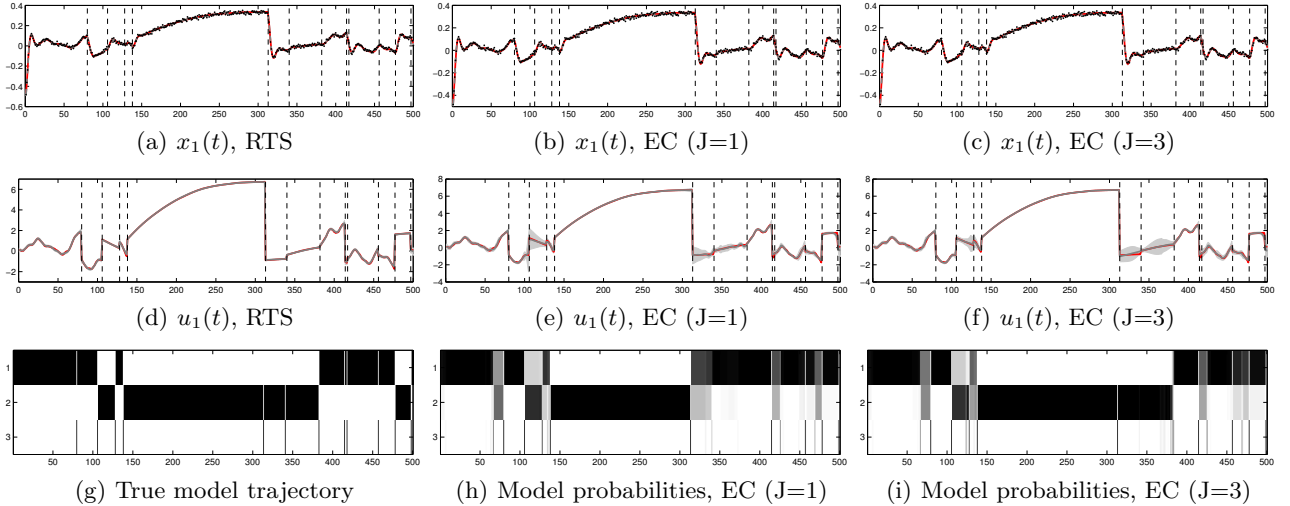

Figure 1: **Illustration of Probabilistic SLFM.** Panels (a)-(c) show the estimates of first output $x_1(t)$ with RTS smoother (with known model trajectory) as well as with EC ($J = 1$ and $J = 3$). Similarly Panels (d)-(f) show the results for latent force $u_1(t)$. Dotted lines denote the switching points, red lines the true process values and dark gray the posterior means. Light gray shade denotes 95% posterior uncertainty. Panel (g) shows the true model sequence and the estimated model probabilities on each time step are shown in (h) and (i), where black denotes 1 and white 0.

### 4.3 GPS Tracking with SLFM

To test how the proposed methods works with real world data, we used it for estimating the switching points of car positioning data. The GPS position data was collected with Indagon's MTT130 positioning terminal, which had Fastrax's IT03 GPS module as the positioning device. The terminal was installed in a conventional passenger car (Volkswagen Golf Variant) and the data was collected using a laptop computer. The GPS antenna was placed on the roof of the car. The test data was collected while driving around on the roads and streets in and around Helsinki, Finland, and it contains stops to traffic lights, crossing turns and various other situations that could be modeled as switches in latent forces. The data is shown as time series and on a two dimensional plane in Panels (a), (b) and (f) of Figure 2.

We modelled the $D = 2$ dimensional GPS data ($T = 6865$) with a switching LFM having $R = 2$ latent forces (Matérn covariances with $\nu = 3/2$) and $L = 2$ possible length-scales. We optimized the hyperparameters of the output model and the length-scales with respect to approximate marginal likelihood given by the ADF ($I = 2$). After learning the parameters we applied EC ($J = 2$). The obtained results are shown in Figure 2. It can be seen that the model is easily able capture the most obvious switching points in the data.

## 5  Conclusions

In the paper [2] it was discussed that the Kalman filtering and smoothing approach has been usually preferred mainly due to computational reasons, but in this article we have shown that LFMs can be equivalently formulated and solved using the state variable approach, which is commonly used in Kalman filtering and smoothing [4, 6]. The state-space view of LFMs actually gives various other advantages in addition to favorable computational efficiency. An example of this is that we can formulate a switching latent force model, in which the switching process can be inferred probabilitistically with methods tailored for switching linear dynamical systems. We have illustrated the performance of the proposed methodology in simulated scenarios and applied it to inferring the switching points in GPS data collected from car movement data in urban environment.

**Acknowledgements**

The authors acknowledge the financial support from Finnish Doctoral Programme in Computational Sciences (FICS), Finnish Foundation for Technology Promotion (TES), the Emil Aaltonen Foundation and Academy of Finland's Centre of Excellence in Com-putational Complex Systems Research. The authors would like to thank Aki Vehtari for helpful discussion and support during the work.

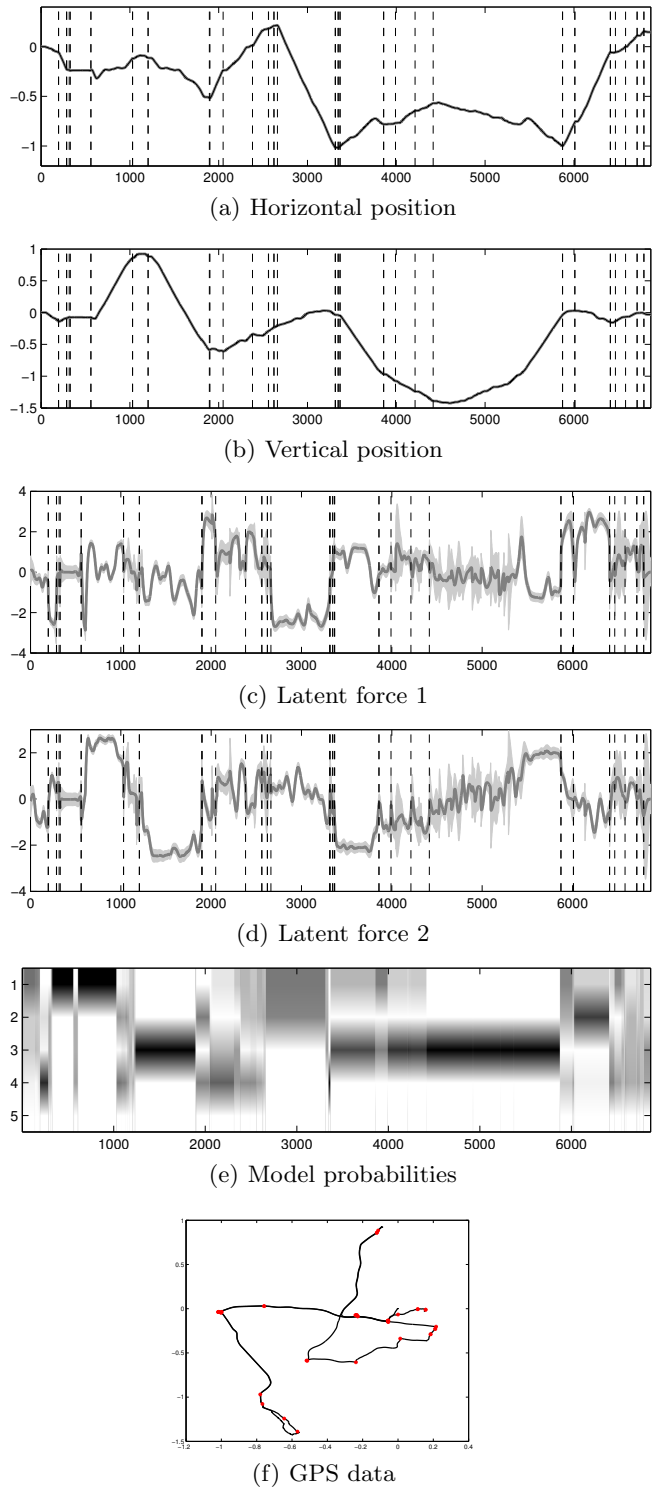

Figure 2: **GPS Tracking with Switching Latent Forces.** Panels (a) and (b) show the horizontal and vertical positions of the car obtained by the GPS. Panels also show the mean estimate produced by EC, which is indistinguishable from the data. One unit in the plots is 10km. Panels (c) and (d) show the estimated latent forces with dark gray denoting the mean estimate and light gray shade the 95% uncertainty. Dotted black bars denote the points, which had over 20% estimated probability of being a switching point. The estimated model probabilities are shown in Panel (e). Panel (f) shows the GPS data on a two dimensional plane together with the estimated switching points (red stars).